\documentclass{article} % For LaTeX2e
\usepackage{conll-2019,times}

\usepackage{url}
\usepackage{booktabs}       % professional-quality tables
\usepackage{amsfonts}       % blackboard math symbols
\usepackage{nicefrac}       % compact symbols for 1/2, etc.
\usepackage{microtype}      % microtypography
\usepackage{amsmath}
\usepackage{color}
\usepackage{multirow}
\usepackage{graphicx}
\usepackage{csquotes}
\usepackage{subfig}
\usepackage{hhline}

\usepackage{enumitem}
\renewcommand{\eqref}[1]{Eq.~(\ref{#1})}

\renewcommand{\cite}{\citep}

\newcommand{\atprior}{AT-Prior} % Alias table baseline
\newcommand{\atext}{AT-Ext} % Alias table baseline extended
\newcommand{\deer}{DEER} % Dual Encoder for Entity Resolution
\newcommand{\bm}{BM25} % BM25 baseline
\newcommand{\demn}{DEER} % dual encoder with mined negatives

\title{Learning Dense Representations for Entity Retrieval}

\author{
Daniel Gillick\Thanks{Equal Contributions} \\
Google Research \\
{\tt \small{ dgillick@google.com}} \\
\And
Sayali Kulkarni\samethanks \\
Google Research \\
{\tt \small{ sayali@google.com}} \\
\And
Larry Lansing\samethanks \\
Google Research \\
{\tt  \small{llansing@google.com}} \\
\And
Alessandro Presta\samethanks \\
Google Research \\
{\tt  \small{apresta@google.com}} \\
\AND
Jason Baldridge\\
Google Research \\
{\tt  \small{jasonbaldridge@google.com}} \\
\And
Eugene Ie\\
Google Research \\
{\tt  \small{eugeneie@google.com}} \\
\And
Diego Garcia-Olano\Thanks{Work done during internship with Google} \\
University of Texas at Austin \\
{\tt  \small{diegoolano@gmail.com }} \\
}

\aclfinalcopy % Uncomment for camera-ready version, but NOT for submission.
\begin{document}

\maketitle

\begin{abstract}
We show that it is feasible to perform entity linking by training a dual encoder (two-tower) model that encodes mentions and entities in the same dense vector space, where candidate entities are retrieved by approximate nearest neighbor search. Unlike prior work, this setup does not rely on an alias table followed by a re-ranker, and is thus the first fully learned entity retrieval model. We show that our dual encoder, trained using only anchor-text links in Wikipedia, outperforms discrete alias table and BM25 baselines, and is competitive with the best comparable results on the standard TACKBP-2010 dataset. In addition, it can retrieve candidates extremely fast, and generalizes well to a new dataset derived from Wikinews. On the modeling side, we demonstrate the dramatic value of an unsupervised negative mining algorithm for this task.
\end{abstract}

\section{Introduction}
\label{sec:intro}

A critical part of understanding natural language is connecting specific textual references to real world entities. In text processing systems, this is the task of \textit{entity resolution}: given a document where certain spans of text have been recognized as \textit{mentions} referring to entities, the goal is to link them to unique entries in a knowledge base (KB), making use of textual context around the mentions as well as information about the entities. (We use the term \textit{mention} to refer to the target span along with its context in the document.)

Real world knowledge bases are large (e.g., English Wikipedia has 5.7M articles), so existing work in entity resolution follows a two-stage approach: a first component nominates candidate entities for a given mention and a second one selects the most likely entity among those candidates. This parallels typical \textit{information retrieval} systems that consist of an index and a re-ranking model. In entity resolution, the index is a table mapping \textit{aliases} (possible names) to entities. Such tables need to be built ahead of time and are typically subject to arbitrary, hard cutoffs, such as only including the thirty most popular entities associated with a particular mention. We show that this configuration can be replaced with a more robust model that represents both entities and mentions in the same vector space. Such a model allows candidate entities to be directly and efficiently retrieved for a mention, using nearest neighbor search.

To see why a retrieval approach is desirable, we need to consider how alias tables are employed in entity resolution systems. In the following example, \textit{Costa} refers to footballer \textit{Jorge Costa}, but the entities associated with that alias in existing Wikipedia text are {\it Costa Coffee}, {\it Paul Costa Jr}, {\it Costa Cruises}, and many others---while excluding the true entity.

\begin{displayquote}
\textit{\textbf{Costa} has not played since being struck by the AC Milan forward...}
\end{displayquote}

\noindent
The alias table could be expanded so that last-name aliases are added for all person entities, but it is impossible to come up with rules covering all scenarios. Consider this harder example:

\begin{displayquote}
\textit{...warned Franco Giordano, secretary of the \textbf{Refoundation Communists} following a coalition meeting late Wednesday...}
\end{displayquote}

\noindent
It takes more sophistication to connect the colloquial expression \textit{Refoundation Communists} to the \textit{Communist Refoundation Party}. Alias tables cannot capture all ways of referring to entities in general, which limits recall.

Alias tables also cannot make systematic use of context. In the \textit{Costa} example, the context (e.g., \textit{AC Milan forward, played}) is necessary to know that this mention does not refer to a company or a psychologist. An alias table is blind to this information and must rely only on prior probabilities of entities given mention spans to manage ambiguity. Even if the correct entity is retrieved, it might have such a low prior that the re-ranking model cannot recover it. A retrieval system with access to both the mention span and its context can significantly improve recall. Furthermore, by pushing the work of the alias table into the model, we avoid manual processing and heuristics required for matching mentions to entities, which are often quite different for each new domain.

This work includes the following contributions:
\begin{itemize}
    \item We define a novel dual encoder architecture for learning entity and mention encodings suitable for retrieval. A key feature of the architecture is that it employs a modular hierarchy of sub-encoders that capture different aspects of mentions and entities.
    \item We describe a simple, fully unsupervised \textit{hard negative mining} strategy that produces massive gains in retrieval performance, compared to using only random negatives.
    \item We show that approximate nearest neighbor search using the learned representations can yield high quality candidate entities very efficiently.
    \item Our model significantly outperforms discrete retrieval baselines like an alias table or BM25, and gives results competitive with the best reported accuracy on the standard TACKBP-2010 dataset.
    \item We provide a qualitative analysis showing that the model integrates contextual information and world knowledge even while simultaneously managing mention-to-title similarity.
\end{itemize}

We acknowledge that most of the components of our work are not novel in and of themselves. Dual encoder architectures have a long history \cite{bromley1994signature,chopra2005learning,yih2011learning}, including for retrieval \cite{Gillick:2018:DualEncoders}. Negative sampling strategies have been employed for many models and applications, e.g. \citet{shrivastava2016training}. Approximate nearest neighbor search is its own sub-field of study \cite{andoni2008near}. Nevertheless, to our knowledge, our work is the first combination of these ideas for entity linking. As a result, we demonstrate the first accurate, robust, and highly efficient system that is actually a viable substitute for standard, more cumbersome two-stage retrieval and re-ranking systems. In contrast with existing literature, which reports multiple seconds to resolve a single mention, we can provide strong retrieval performance across all 5.7 million Wikipedia entities in around 3ms per mention.
\section{Related work}
\label{sec:related}

Most recent work on entity resolution has focused on training neural network models for the candidate re-ranking stage \cite{francislandau-durrett-klein:2016:N16-1,eshel-EtAl:2017:CoNLL,DBLP:journals/tacl/YamadaSTT17,DBLP:conf/emnlp/GuptaSR17,sil:etal:aaai2018}. In general, this work explores useful context features and novel architectures for combining mention-side and entity-side features. Extensions include joint resolution over all entities in a document \cite{Ratinov:2011:LGA:2002472.2002642,globerson2016,DBLP:journals/corr/GaneaH17}, joint modeling with related tasks like textual similarity \cite{DBLP:journals/corr/YamadaS0T17,barrena2018learning} and cross-lingual modeling \cite{sil:etal:aaai2018}, for example.

By contrast, since we are using a two-tower or dual encoder architecture \cite{Gillick:2018:DualEncoders,Serban:2018:DualEncoders}, our model cannot use any kind of attention over both mentions and entities at once, nor feature-wise comparisons as done by \citet{francislandau-durrett-klein:2016:N16-1}. This is a fairly severe constraint -- for example, we cannot directly compare the mention span to the entity title -- but it permits retrieval with nearest neighbor search for the \textit{entire} context against a single, all encompassing representation for each entity.

\section{Data}
\label{sec:data}

As a whole, the entity linking research space is fairly fragmented, including many task variants that make fair comparisons difficult. Some tasks include named entity recognition (mention span prediction) as well as entity disambiguation, while others are concerned only with disambiguation (the former is often referred to as \textit{end-to-end}. Some tasks include the problem of predicting a \textit{NIL} label for mentions that do not correspond to any entity in the KB, while others ignore such cases. Still other tasks focus on \textit{named} or proper noun mentions, while others include disambiguation of concepts. These variations and the resulting fragmentation of evaluation is discussed at length by \citet{ling2015design} and \citet{hachey2013evaluating}, and partially addressed by attempts to consolidate datasets \cite{cornolti2013framework} and metrics \cite{usbeck2015gerbil}.

Since our primary goal is to demonstrate the viability of our unified modeling approach for entity retrieval, we choose to focus on just the disambiguation task, ignoring NIL mentions, where our set of entity candidates includes every entry in the English Wikipedia.

In addition, some tasks include relevant training data, which allows a model trained on Wikipedia (for example) to be tuned to the target domain. We save this fine-tuning for future work.

\begin{figure*}
  \centering
  \includegraphics[width=\textwidth]{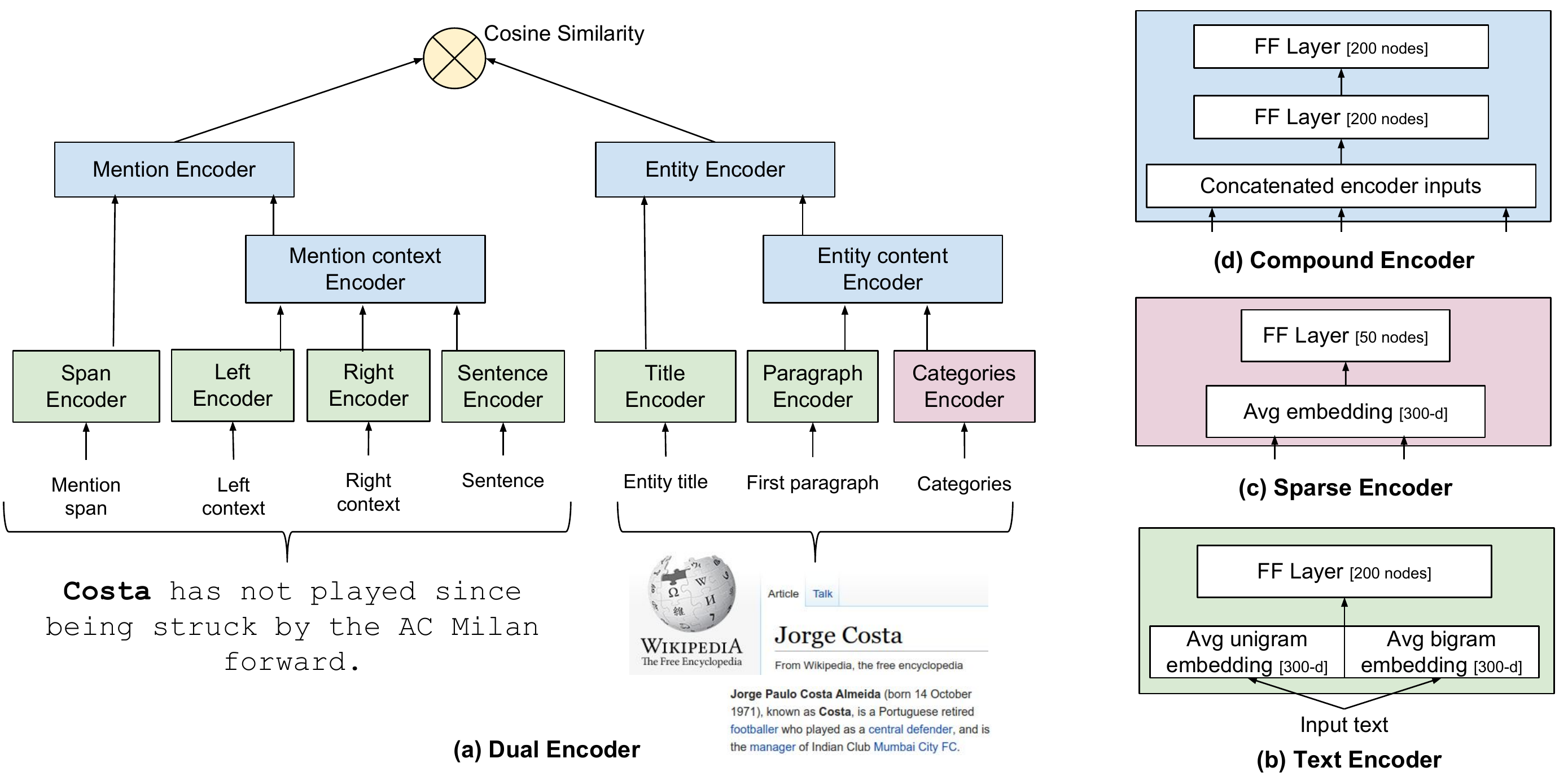}
  \caption{Architecture of the dual encoder model for retrieval (a). Common component architectures are shown for (b) text input, (c) sparse ID input, and (d) compound input joining multiple encoder outputs. Note that all text encoders share a common set of embeddings.\label{fig:architecture}}
\end{figure*}

\paragraph{Training data}

Wikipedia is an ideal resource for training entity resolution systems because many mentions are resolved via internal hyperlinks (the mention span is the anchor text). We use the 2018-10-22 English Wikipedia dump, which includes 5.7M entities and 112.7M linked mentions (labeled examples). We partition this dataset into 99.9\% for training and the remainder for model selection.

Since Wikipedia is a constantly growing an evolving resource, the particular version used can significantly impact entity linking results. For example, when the TACKBP-2010 evaluation dataset was published, Wikipedia included around 3M entities, so the number of retrieval candidates has increased by nearly two times. While this does mean new contexts are seen for many entities, it also means that retrieval gets more difficult over time. This is another factor that makes fair comparisons challenging.

\paragraph{Evaluation data}
There are a number of annotated datasets available for evaluating entity linking systems. Given the choices discussed above, the \textbf{TACKBP-2010} dataset\footnote{\url{https://tac.nist.gov/}} is the most widely used evaluation that matches our constraints and allows us to compare to a reasonable variety of prior work. It includes 1020 annotated mention/entity pairs derived from 1013 original news and web documents. While there is a related development set associated with this evaluation set, we do not use it for any fine-tuning, as explained above.

To further validate our results, we also include a new evaluation set called \textbf{Wikinews}, which includes news pages from Wikinews\footnote{\url{https://en.wikinews.org}} in English for the year 2018. It includes 2263 annotated mention/entity pairs derived from 1801 original documents. Because we pulled these documents at the same time as the Wikipedia dump, the entity annotations are consistent with our training set and have not been subject to the kind of gradual rot that befalls older evaluation data as the updated KB diverges from the annotations. This data is available here: \url{https://github.com/google-research/google-research/tree/master/dense_representations_for_entity_retrieval/}.

\section{Entity retrieval model}
\label{sec:model}

We use nearest neighbor search to retrieve entities based on a mention in context, after learning dense, fixed-length vector representations of each.

\subsection{Dual Encoder model}

The \textit{dual encoder} is a two-tower architecture suitable for retrieval \cite{Gillick:2018:DualEncoders}. It has one network structure for encoding mentions (including their contexts), a second for encoding entities (including KB features), and a cosine function to compute similarity between representations (which must be the same dimension). A key property of this architecture is that there is no direct interaction between the encoders on each side. This enables efficient retrieval, but constrains the set of allowable network structures. The dual encoder learns a mention encoder $\phi$ and an entity encoder $\psi$, where the score of a mention-entity pair $(m, e)$ defined as:
\begin{equation}
    s(m, e) = \cos(\phi(m), \psi(e))
\end{equation}

Figure~\ref{fig:architecture} shows the full model architecture and the feature inputs to each encoder. We use a \textit{compound encoder} (Figure \ref{fig:architecture}d) to add useful sub-structure to each tower. The mention-side encoder first combines the context features, and then combines the result with the mention span encoding. Similarly, the entity-side encoder first combines the entity paragraph and categories, and then combines the result with the entity title encoding. 

The mention encoder uses four text features to capture both the span text and the textual context surrounding it. The context features include the five tokens immediately to the left and right of the span. In the sentence feature, the mention span is replaced by a special symbol.

The entity encoder uses the entity's title and the first paragraph of its Wikipedia page as text features. It additionally incorporates the unedited user-specified categories associated with the entity. We do not use the entity IDs as features so that the model generalizes more easily to new entities unseen at training time. In fact, more than 1M candidate entities available at retrieval time have no associated training examples, but this architecture allows these to be encoded using their feature representations.

A shared embedding look-up is used for all text features (Figure \ref{fig:architecture}b). Specifically, we embed all unigrams and bigrams to get 300-dimensional averaged unigram embeddings and 300-dimensional averaged bigram embeddings for each text feature. Unigram embeddings are initialized from GloVe vectors \cite{Pennington14glove:global}, and we use 5M hash buckets for out-of-vocabulary unigrams and bigrams\cite{ganchev2008small}. These averaged embeddings are concatenated and then passed through a feed-forward layer. For the category features, each entity category name is treated as a sparse input, and the embeddings for all categories for an entity are averaged to produce a 300-dimensional representation, which in turn is passed through a feed-forward layer (Figure \ref{fig:architecture}c).

Our experiments show that this architecture is highly effective for both retrieval and resolution. Nevertheless, we expect that additional modeling ideas will further improve performance, especially for resolution. Recent work such as \citet{DurrettKlein2014} has shown improvements derived from better, longer-range, context features; similarly, there are many more potentially useful KB-derived features. More complex encoder architectures that use some form of attention over the input tokens and features could also be beneficial. 

\subsection{Training}

Our training data is described in Section \ref{sec:data}. The inputs to the entity encoder are constructed from the true entity referred by the landing page. The inputs to the mention encoder are constructed from the source page, using the mention span and surrounding context.

These pairs constitute only positive examples, so we use \textit{in-batch random negatives} \cite{henderson2017efficient,Gillick:2018:DualEncoders}: for each mention-entity pair in a training batch, the other entities in the batch are used as negatives. Computationally, this amounts to building the all-pairs similarity matrix for all mentions and entities in a batch. We optimize softmax loss on each row of the matrix, so that the model is trained to maximize the score of the correct entity with respect to random entities. This is a version of the sampled softmax \cite{jozefowicz2016exploring}, which we use in place of the full softmax because the normalization term is intractable to compute over all 5.7M entities.

The softmax loss is not directly applied to the raw cosine similarities. Instead, a scalar multiplier $a$ is learned to map the similarities (in the range $[-1,1]$) to unbounded \textit{logits}. For each training pair $(m_i, e_i)$ in a batch of $B$ pairs, the loss is computed as:
\begin{equation}
    L(m_i, e_i) = -f(m_i, e_i) + \log\sum_{j = 1}^{B} \exp(f(m_i, e_j))
\end{equation}
where
\begin{equation}
     f(m_i, e_j) = a \cdot s(m_i, e_j)
\end{equation}

We track in-batch recall@1 (accuracy) on the held out set during training. For each instance, the model gets a score of 1 if the correct entity is ranked above all in-batch random negatives, 0 otherwise. We stop training after the metric flattens out (about 40M steps).

For all experiments, we use a batch size of 100, standard SGD with Momentum of 0.9 and a fixed learning rate 0.01. 

Our aim here is to demonstrate a pure retrieval system, so we train our models solely from Wikipedia and refrain from tuning them explicitly on in-domain documents from the evaluation tasks.

\subsection{Hard negative mining}
\label{subsec:hardnegs}

Random negatives alone are not enough to train an accurate entity resolution model because scoring the correct entity above random alternatives can typically be achieved just by comparing the mention text and entity title. More challenging negative examples must be introduced to force the model to exploit context. This strategy is somewhat akin to Importance Sampling \cite{bengio2003quick}, for example.

After learning an initial model using random negatives, we identify hard negatives via the following steps:

\begin{enumerate}
    \item Encode all mentions and entities found in training pairs using the current model.
    \item For each mention, retrieve the most similar 10 entities (i.e., its nearest neighbors).
    \item Select all entities that are ranked above the correct one for the mention as negative examples.
\end{enumerate}

\noindent
This yields new negative mention/entity pairs for which the model assigns a high score. It crucially relies on the fact that there is just one correct entity, unlike other tasks that consider general similarity or relatedness (and which are well served by random negatives).
For example, negatives mined in this way for paraphrasing or image captioning tasks could actually turn out to be positives that were not explicitly labeled in the data. It is precisely because the distribution over candidate entities that match a contextualized mention tends to have such low entropy that makes negative mining such a good fit for this task.

After merging these with the original positive pairs to construct a classification task, we resume training the initial dual encoder model using logistic loss on this new set of pairs. To retain good performance on random negatives, the new task is mixed with the original softmax task in a multi-task learning setup in which the two loss functions are combined with equal weight and optimized together. For a pair $(m, e)$ with label $y \in \{0,1\}$, the hard negative loss is defined as:

\begin{equation}
\begin{aligned}
    L_h(m, e; y) = & -y \cdot \log{f(m, e)} \\
                   & -(1-y) \cdot \log(1-f(m, e))
\end{aligned}
\end{equation}
where
\begin{equation}
    f(m, e) = g(a_h \cdot s(m, e) + b_h)
\end{equation}

\noindent
Here, $g(x) = 1/(1+e^{-x})$ is the logistic function, and $a_h$ and $b_h$ are learned scalar parameters to transform the cosine similarity into a logit.\footnote{The additive parameter is only needed for the logistic loss component, as the softmax function is invariant to translation.}

For the hard negatives task, we track \textit{Area Under the ROC curve} (AUC) on a held out set of pairs. We stop training when the average of the evaluation metrics for the two tasks stabilizes (about 5M steps).

Finally, we iteratively apply this negative mining procedure. For each round, we mine negatives from the current model as described above and then append the new hard examples to the classification task. Thus each subsequent round of negative mining adds fewer and fewer new examples, which yields a stable and naturally convergent process.
As we show in our experiments, iterative hard negative mining produces large performance gains.

\subsection{Inference}
Once the model is trained, we use the entity encoder to pre-compute encodings for all candidate entities (including those that do not occur in training). At prediction time, mentions are encoded by the mention encoder and entities are retrieved based on their cosine similarity. Since our focus is on model training, we use brute-force search in our evaluation. However, for online settings and larger knowledge bases, an approximate search algorithm is required. In Section~\ref{subsec:approx} we show that, when using approximate search, the system retains its strong performance while obtaining a nearly 100x speedup on an already fast retrieval.
\section{Experiments}
\label{sec:experiments}

\subsection{Evaluation setup}

We demonstrate our model performance as compared to a baseline alias table. As is standard, it is built by counting all (mention span, entity) pairs in the training data. The counts are used to estimate prior probabilities $P(e|m)$ of an entity given a mention span (alias); for each entity, the aliases are ordered according to these priors and limited to the top 100 candidates.

\begin{equation}
P(e|m) = \frac{count(e,m)}{\sum\limits_{e' \in E}{count(e',m)}}
\end{equation}

\noindent
Here, $count(e,m)$ is the number of occurrences of mention $m$ linked to entity $e$, and $E$ is the set of all entities appearing in the data. 

Since alias table construction is often extended with various heuristics, we also include a variant that includes unigrams and bigrams of the mention text as additional aliases. This can help when the entities (specifically person names) are referenced as last/first name at inference time.

Finally, since we are primarily concerned with demonstrating performance of a retrieval system (as opposed to a re-ranking system) or a combination of the two, we include results using the standard BM25 retrieval algorithm (the \textit{Gensim} implementation\footnote{\url{https://radimrehurek.com/gensim/summarization/bm25.html}}). We found that indexing each entity using its title gave much better results than indexing with the first paragraph text (or the full document text).

We measure recall@k (R@k), defined as the proportion of instances where the true entity is in the top $k$ retrieved items. We report R@1 (accuracy of the top retrieved result), which is standard for TAC/KBP-2010, as well R@100, which better captures overall retrieval performance.

We refer to the models with these abbreviations:

\begin{itemize}
\item \textbf{\atprior}: The alias table ordered by $P(e|m)$.
\item \textbf{\atext}: The heuristically extended alias table.
\item \textbf{\bm}: The BM25 retrieval algorithm, where each entity is indexed using its title.
\item \textbf{\deer}: Our Dual Encoder for Entity Resolution, as described in section \ref{sec:model}.
\end{itemize}

\subsection{Results}

% New table with TAC 2010 comparison with prior work
\begin{table}
\centering
\begin{tabular}{c|c|c}
     \textbf{System} & \textbf{R@1} & \textbf{Entities}\\
    \hline
    AT-Prior & 71.9 & 5.7M\\
    AT-Ext   & 73.3 & 5.7M \\
    \hline
     \citet{chisholm2015entity} & 80.7 & 800K \\
     \citet{he-EtAl:2013:Short} & 81.0 & 1.5M \\
     \citet{sun2015modeling} & 83.9 & 818K \\
     \citet{yamada2016joint} & 85.2 & 5.0M \\
     \citet{nie:etal:aaai2018} & 86.4 & 5.0M \\
     \citet{barrena2018learning} & 87.3 & 523K \\
     \textbf{\deer \: (this work)} & 87.0 & 5.7M\\
\end{tabular}
\caption{Comparison of relevant TACKBP-2010 results using Recall@1 (accuracy). While we cannot control the candidate entity set sizes, we attempt to approximate them here. \label{tab:tac2010}}
\end{table}

Table \ref{tab:tac2010} provides a comparison  against the most relevant related work. While there are some reported improvements due to collective (\textit{global}) resolution of all mentions in a document (\citet{globerson2016} report 87.2\% and \citet{nie:etal:aaai2018} report 89.1\%), we limit comparisons to \textit{local} resolution. We also limit comparisons to systems that ignore NIL mentions (referred to as \textit{in-KB} accuracy), so all those reported in the table evaluate precisely the same set of mentions. As noted earlier, we cannot control the candidate sets used in each of these experiments, and we are at some disadvantage given our larger set of candidates.

\paragraph{Retrieval performance} Table \ref{tab:r@100} provides the percent of mentions for which the correct entity is found in the top 100 retrieved results, using the different baselines and the \deer\ model. The learned representations deliver superior performance and do not require special handling for unigrams versus bigram lookups, counts for entity prominence, and so on.

\paragraph{Resolution performance}
To put \deer's architecture and performance in context, we compare it with prior work in some more detail here.

\citet{he-EtAl:2013:Short} use a dual encoder setup to train a re-ranker, but start with unsupervised training to build representations of contexts and entities using Denoising Autoencoders. They use an alias table for candidate generation, and then train a ranking model using mention-specific batching to obtain hard in-batch negatives. Our results suggest that the autoencoder pretraining is not necessary and that our unsupervised negative mining can outperform heuristic selection of negatives.

\citet{sun2015modeling} also use a dual encoder that has similar structure to ours. Like \citet{he-EtAl:2013:Short},  they use it to score entries from an alias table rather than directly for retrieval. Their mention encoder is a considerably more complex combination of mention and context rather than the simple compounding strategy in our architecture. Their alias table method not only maps mentions to entities, but also uses additional filters to reduce the set of candidate entities based on words in the mention's context. They report that this method has a recall of 91.2\% on TACKBP 2010, while our direct retrieval setup gives Recall@100 of 96.3\% (see Table \ref{tab:r@100}). They train their representations for each true mention-entity pair against a single random negative entity for the mention, whereas our method takes advantage of the entire batch of random negatives as well further refinement through hard negative mining.

% New table with R@100 comparison with baselines on TAC and Wikinews
\begin{table}
\centering
\begin{tabular}{c|c|c}
\textbf{System} & \textbf{TACKBP-2010} & \textbf{Wikinews} \\
\hline
\atprior & 89.5 & 93.8 \\
\atext   & 91.7 & 94.0 \\
\bm & 68.9 & 83.2 \\
\textbf{\deer} & \textbf{96.3} & \textbf{97.9} \\
\end{tabular}
\caption{Retrieval evaluation comparison for TACKBP-2010 and Wikinews using Recall@100. \label{tab:r@100}}
\end{table}

\citet{yamada2016joint} use an alias table derived from the December 2014 Wikipedia dump, restricted to the fifty most popular entities per mention. They tune their model on the TACKBP 2010 training set. Architecturally, they include features that capture the alias table priors and string similarities, both of which are not feasible in a dual encoder configuration that precludes direct comparison between mention- and entity-side features. \deer's better results indicate that learned representations of mentions and entities can be powerful enough for entity retrieval even without any cross-attention.

\citet{nie:etal:aaai2018} define a complex model that uses both entity type information and attention between the mention string and the entity description. To augment the small 1500 example training data in TACKBP, they also collected 55k mentions found in Wikipedia that were active in TACKBP 2010 to train this model. \deer\ is simply trained over all entities in Wikipedia and uses no cross-attention or explicit type information, yet delivers better resolution performance.

Most standard entity linking models build a single ranking model on top of the candidate set provided by an alias table. \citet{barrena2018learning} instead train 523k mention-specific deep classifiers---effectively treating entity linking as a special form of word sense disambiguation. They do this by pre-training a single LSTM that predicts among 248k mentions, and then the parameters of this model are used to warm start each of the 523k mention-specific models. In doing so, they learn an effective context encoding, and can then fine-tune each mention model to discriminate among the small set of popular candidate entities for the mention (their alias table uses a cutoff of the thirty most popular entities for each mention). \deer\, in contrast, has a single mention encoder that is simple and fast, and performs nearly equivalently while retrieving from a much larger set of entities.

\begin{table}[]
    \centering
    \begin{tabular}{c|cc}
         \textbf{Method} & \textbf{\shortstack{Mean \\ search time (ms)}} & \textbf{\shortstack{Wikinews \\ R@100}}  \\
         \hline
         Brute force & 291.9 & 97.88  \\
         AH & 22.6 & 97.22 \\
         AH+Tree & 3.3 & 94.73 \\
    \end{tabular}
    \caption{Comparison of nearest-neighbor search methods using the \deer\ model. The benchmark was conducted on a single machine. AH indicates quantization-based asymmetric hashing; AH+Tree adds an initial tree search to further reduce the search space.\label{tab:approx}}
\end{table}

\paragraph{Approximate search}
\label{subsec:approx}

Tables \ref{tab:tac2010} and \ref{tab:r@100} report performance using brute force nearest-neighbor search. That is, we score each mention against all 5.7M entities to get the top $k$ neighbors. However, a major motivation for using a single-stage retrieval model is that it can allow scaling to much larger knowledge bases by reducing retrieval time via approximate search.

To estimate performance in a real-world setting, we repeat the evaluation of \deer\ using the quantization-based approaches described by \citet{guo2016quantization}. Table~\ref{tab:approx} shows the trade-off between search time and recall on Wikinews. Compared to brute force, search time can be reduced by an order of magnitude with a small loss in R@100, or by two orders of magnitude while losing less than 3 points. This is crucial for scaling the approach to even larger KBs and supporting the latency requirements of real-world applications.

\paragraph{Impact of hard negative mining} 

\begin{figure}
    \centering
    {\includegraphics[width=7cm]{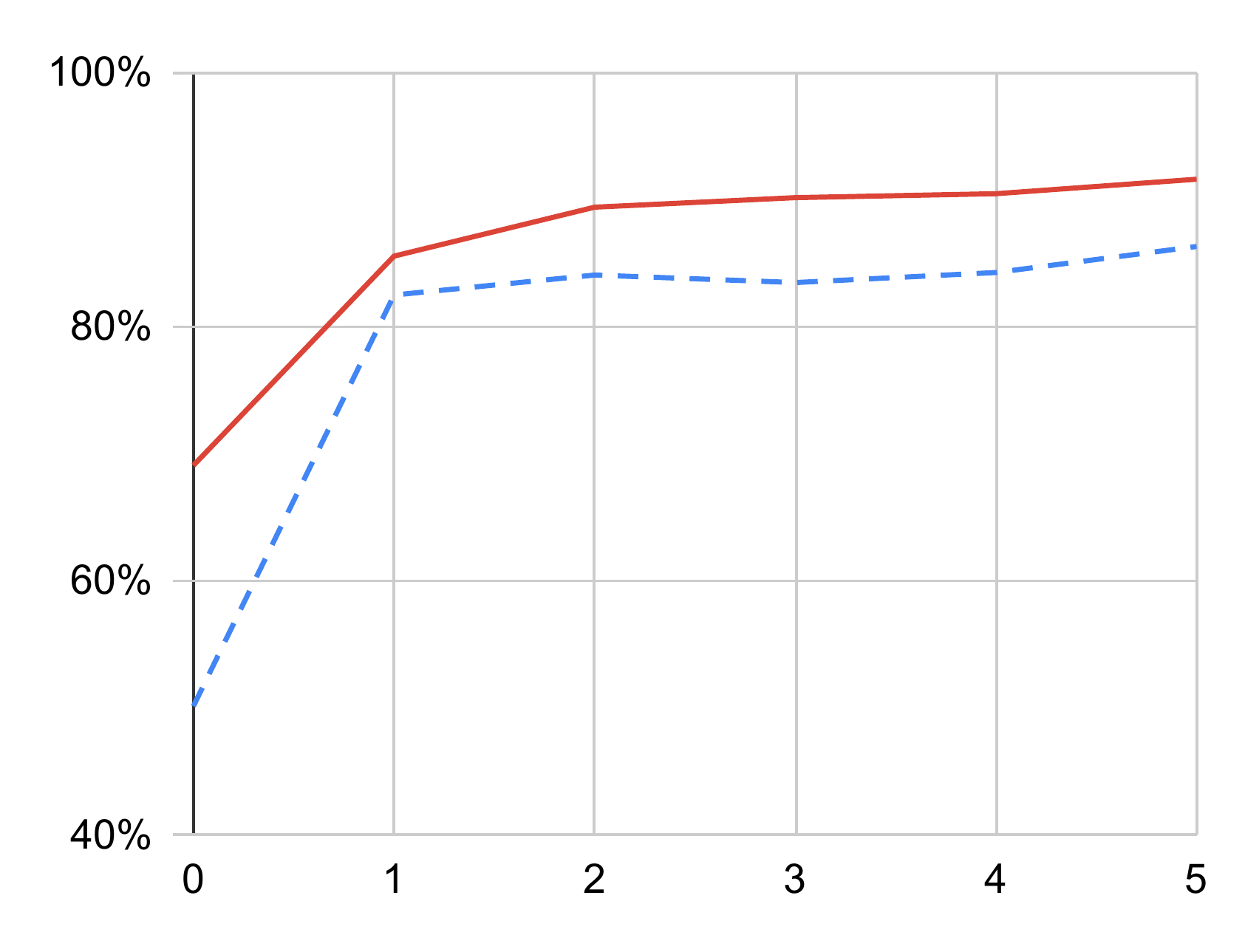}}
    \caption{Recall@1 improvement for successive iterations of hard negative mining for Wikinews (solid) and TACKBP-2010 (dashed).}
    \label{fig:iterative_hn}
\end{figure}

Figure~\ref{fig:iterative_hn} shows the improvement in Recall@1 from each round of hard negative mining. The first iteration gives a large improvement over the initial round of training with only random negatives. Successive iterations yield further gains, eventually flattening out. Our strategy of appending each new set of hard negatives to the previously mined ones means that each new set has proportionately less influence---this trades off some opportunity for improving the model in favor of stability.

\subsection{Qualitative analysis}

Here, we show a variety of examples to elucidate how DEER effectively models context, and to provide intuition for the learned entity representations.

First, we compare entities retrieved by DEER with those retrieved by the alias table baseline. Table~\ref{tab:retrieval_examples} shows some instances where the alias table does not contain the correct entity for a given mention text (in the top 100 neighbors) or it fails to return any entity at all. In all of these cases, it is clear that context is essential. While a scoring stage is intended to leverage context, it is limited by the set of retrieved entities; our approach uses context directly.

\begin{table}
\centering
\begin{tabular}{p{0.1\textwidth}|p{0.33\textwidth}}
\textbf{Entity} & \textbf{Nearest neighbors} \\ \hline
Jorge Costa & Jos\'e Alberto Costa, Eduardo Costa, Peter Shilton, Rui Costa, Nuno Gomes, Ricardo Costa (Portuguese footballer), Andr\'e Gomes, Bruno Ribeiro, Diego Costa  \\ \hline
Costa Cruises & MSC Cruises, P\&O Cruises, Princess Cruises, Island Cruises, AIDA Cruises, Silversea Cruises, Carnival Corporation \& plc, Costa Concordia, Celebrity Cruises \\ \hline
Arctic sea ice decline & Arctic ice pack, Measurement of sea ice, Arctic geoengineering, Arctic sea ice ecology and history, Climate Change Science Program, Abrupt climate change, Sea ice thickness, Antarctic sea ice, Marine ice sheet instability \\ \hline
Pink Floyd & Led Zeppelin, The Who, Duran Duran, Syd Barrett, The Velvet Underground, Eddie Floyd, The Beatles, The Australian Pink Floyd Show, Roger Waters \\
\end{tabular}
\caption{Nearest neighbors retrieved by \demn\ for a sample of entities.}
\label{tab:nearest_neighbors}
\end{table}

For example, mentions like \textit{Costa} and \textit{Justin} are linked to the correct entities in the alias table, but with such low prior probability that we would need to retrieve far more than the top 100 entities to consider them. At the other extreme, mentions like \textit{Refoundation Communists} and \textit{European EADS} are missed by the baseline because they don't have direct string matches in the alias table. Additional extensions to our alias table allowing token re-ordering could help catch the former (though this might reduce precision too much), but it's unlikely that any alias table could connect \textit{European EADS} with \textit{Airbus} in the absence of an explicit anchor-text link. This example helps highlight how a fully learned retrieval model can generalize to new data.

Second, Table~\ref{tab:context_effect_on_nn} shows more directly how modifying the context around a mention span changes the retrieved entities. For example, the model correctly differentiates between Phoenix the city, Phoenix the band, and Phoenix in mythology based on the sentence surrounding an identical mention span.

\begin{table*}
\centering
\small
\begin{tabular}{p{0.4\textwidth}|p{0.27\textwidth}|p{0.27\textwidth}}
\textbf{Mention} & \textbf{Baseline predictions} & \textbf{Model predictions} \\ \hline
\textbf{Costa} has not played since being struck by the AC Milan forward & Costa Coffee, Paul Costa Jr, Comstock-Needham system, Costa Cruises, Achille Costa & Ricardo Costa (Portuguese footballer), Fernando Torres, Pedro (footballer born 1987), \textbf{Jorge Costa} \\ \hline
Australia beat West Indies by five wickets in a \textbf{World Series} limited overs match & World Series, ATP International Series, 2010 World Series & \textbf{World Series Cricket}, The University Match (cricket), Australian Tri-Series \\ \hline
\textbf{Justin} made his second straight start for Harbaugh, who has a knee injury & Justin (historian), Justin Martyr, Justin (consul 540) & \textbf{Paul Justin}, Joe Montana, Dale Steyn \\ \hline 
plays for the Cape Town-based \textbf{Cobras} franchise & Cobra, AC Cobra, Indian Cobra & Cobra, Snake, \textbf{Cape Cobras} \\ \hline
\textbf{OSI} reports profit on overseas cancer drug sales & Open Systems Interconnection, Open Source Initiative, OSI (band) & Health Insurance Portability and Accountability Act, \textbf{OSI Pharmaceuticals}, James L. Jones \\ \hline
EVN imposed \textbf{rotating power cuts} earlier this year as the worst drought in a century dropped water levels & no matches & Cogeneration, \textbf{Power outage}, Scram \\ \hline 
warned Franco Giordano, secretary of the \textbf{Refoundation Communists} following a coalition meeting late Wednesday & no matches & League of Communists of Yugoslavia, \textbf{Communist Refoundation Party}, Communist Party of Spain \\ \hline
The \textbf{European EADS} consortium, which makes the Eurofighter Typhoon, said it was not comfortable with the NATO-member countries' bidding process & no matches & \textbf{Airbus}, Airbus A400M Atlas, NATO \\ \hline
such as the record California wildfires, high temperature extremes, retreating glaciers, and melting snow cover, the \textbf{decline of sea ice}, rising sea levels with increasing ocean acidification and coastal flooding & no matches & Moskstraumen, \textbf{Arctic sea ice decline}, Glacier, Sea ice \\
\end{tabular}
\caption{Examples of test mentions that require making use of context, where the alias table does not retrieve the correct entity. We show the top entities returned by both systems, with the correct entity in bold.}
\label{tab:retrieval_examples}
\end{table*}

\begin{table*}
\centering
\small
\begin{tabular}{p{0.65\textwidth}|p{0.3\textwidth}}
\textbf{Mention} & \textbf{Model predictions} \\ \hline
From 1996, \textbf{Cobra} was brewed under contract by Charles Wells Ltd and experienced strong growth in sales for the next ten years. & The Cobra Group, \textbf{Cobra Beer}, Cobra (Tivoli Friheden) \\ \hline
Guys fondly remembered \textbf{Cobra} - the band from Memphis featuring Jimi Jamison and Mandy Meyer who released one Album - Frist Strike - before the Band split! & \textbf{Cobra (American band)}, Wadsworth Jarrell, Cobra Records, Cobra (Japanese band) \\ \hhline{==}
Since the late 18th century, \textbf{Paris} has been famous for its restaurants and haute cuisine, food meticulously prepared and artfully presented. & \textbf{Paris}, Nice, Bucharest \\ \hline
Kim Kardashian may be a household name now, but that wasn’t always the case - and it may all be because of pal \textbf{Paris}. & Paris, \textbf{Paris Hilton}, Paris syndrome \\ \hline 
Rory and \textbf{Paris} are the only two people on Gilmore Girls who share the same goals. & Paris, Paris (mythology), \textbf{Paris Geller} \\ \hhline{==}
\textbf{Texas} was finally annexed when the expansionist James K. Polk won the election of 1844 who ordered General Zachary Taylor south to the Rio Grande on January 13, 1846. & \textbf{Texas}, Texas annexation, Texas in the American Civil War \\ \hline
Fronted by Sharleen Spiteri, \textbf{Texas} have released eight studio albums and are known for songs such as 'I Don't Want a Lover', 'Say What You Want', 'Summer Son' and 'Inner Smile' & \textbf{Texas (band)}, Texas, Tich (singer)  \\ \hhline{==}
There is an amazing piece of historic architecture set in downtown \textbf{Phoenix} that was build in 1929 in the Spanish Baroque style and features intricate murals and moldings. & \textbf{Phoenix, Arizona}, Prescott, Arizona \\ \hline
\textbf{Phoenix} once again played another late night show, now they have Late Night with Jimmy Fallon where they played a great rendition of 'Lisztomania' & \textbf{Phoenix (band)}, Joaquin Phoenix, Phoenix, Arizona \\ \hline
According to Greek mythology, the \textbf{Phoenix} lived in Arabia next to a well where the Greek sun-god Apollo stopped his chariot in order to listen to its song. & Phoenix (mythology), Phoenix (son of Amyntor), \textbf{Phoenix (son of Agenor)} \\ \hline
\end{tabular}
\caption{Changing the context around a mention span changes the mention encoding, and thus the set of retrieved neighbors.}
\label{tab:context_effect_on_nn}
\end{table*}

Third, since our model produces encodings for all 5.7M entities, we can retrieve nearest neighbors for any entity. Some examples are shown in Table~\ref{tab:nearest_neighbors}. The model tends to prefer related entities of the same type, and often ones that share portions of their names, probably because entity titles are so important to linking with mentions. The nearest neighbors for \textit{Jorge Costa}, our running example, include a variety of retired Portuguese football players, many of whom have \textit{Costa} in their names.

\begin{figure*}
    \centering
    \includegraphics[width=14cm,clip]{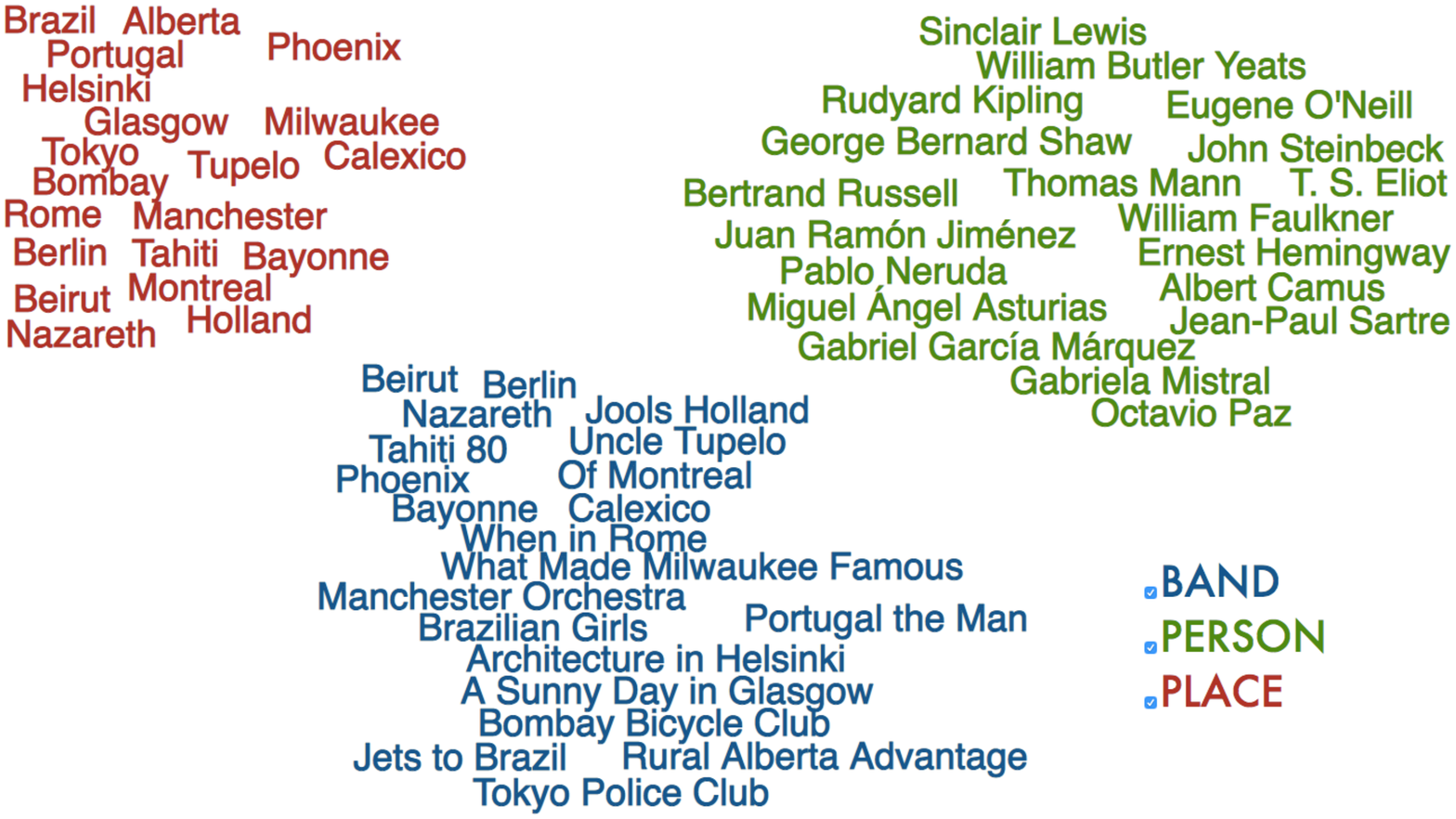}
    \vspace*{-12mm}
    \caption{A 2D projection of cities, bands, and people embeddings (using t-SNE), color coded by their category.}
    \label{fig:citybands}
\end{figure*}

Finally, Figure~\ref{fig:citybands} is a t-SNE projection of the entity encodings for a selection of cities, bands, and people (nobel literature winners). The cities and bands were chosen to have high word overlap, e.g. \textit{Montreal (city)} and \textit{Of Montreal (band)}, to demonstrate how our entity embeddings differ from standard word embeddings. Note also the sub-clusters that form within each type cluster. Latin American authors cluster together, as do the Existentialists; the cities have some geographical proximity, though Brazil and Portugal are neighbors, presumably because of shared language and culture.

\section{Conclusion}
\label{sec:conclusion}

%Recent work on entity resolution \cite{DBLP:journals/corr/GaneaH17} has used network layers with attention over features of both mention and entity---using such models as rerankers over our retrieval results will likely improve R@1 performance (which is not our primary goal in this work).

Our results with DEER show that a single-stage retrieval approach for entities from mentions is highly effective: without any domain-specific tuning, it performs at least as well as the best comparable two-stage systems. While our bag-of-ngrams encoders provided a strong proof of concept, we can almost certainly improve results with more sophisticated encoders, using a BERT architecture \cite{devlin2018bert}, for example. Further, by virtue of approximate search techniques, it can be used for very fast retrieval, and is likely to scale reasonably to much larger knowledge bases.

We also note that the dual encoder approach allows for interesting extensions beyond traditional entity linking. For example, the context encodings provide a natural model for building entity expectations during text processing, such that entities relevant to the context can be retrieved and used for reference resolution as a document is processed incrementally. We expect this will be useful for collective entity resolution as well as modeling coherence.

Finally, while we focus on training with English Wikipedia, \citet{sil:etal:aaai2018} show that using cross-lingual datasets can help to refine the context information more effectively. Since English constitutes only a fraction of the total Wikipedia, and entity IDs are (mostly) language-independent, there is great opportunity to extend this work to far more training examples across far more languages.
%\newpage

%\pagebreak

\section*{Acknowledgements}
The authors would like to thank Ming-Wei Chang, Jan Botha, Slav Petrov, and the anonymous reviewers for their helpful comments.

\nocite{zhang2018generative}
\nocite{globerson2016}
\nocite{DBLP:journals/corr/abs-1708-00781}
\nocite{DBLP:journals/corr/GaneaH17}

\bibliography{mybib}
\bibliographystyle{acl_natbib}

\end{document}